\documentclass[sn-apa]{sn-jnl}

\usepackage{graphicx}
\usepackage{multirow}
\usepackage{array}
\usepackage{url}
\usepackage[caption=false]{subfig}
\usepackage{xcolor}

\newcolumntype{C}[1]{>{\centering\arraybackslash}m{#1}}
\newcolumntype{L}[1]{>{\raggedleft\arraybackslash}m{#1}}
\newcolumntype{R}[1]{>{\raggedright\arraybackslash}m{#1}}
\newcommand\tstrut{\rule{0pt}{2.4ex}}
\newcommand\bstrut{\rule[-1.0ex]{0pt}{0pt}}

\jyear{2021}%
\theoremstyle{thmstyleone}%
\theoremstyle{thmstyletwo}%

\theoremstyle{thmstylethree}%

\begin{document}
\hbadness=1000000000
\vbadness=1000000000

\renewcommand{\thefootnote}{\fnsymbol{footnote}} 

\title[Going Deeper into Recognizing Actions in Dark Environments]{Going Deeper into Recognizing Actions in Dark Environments: A Comprehensive Benchmark Study}


\author[1,2]{\fnm{Yuecong} \sur{Xu}}\email{xuyu0014@e.ntu.edu.sg}
\equalcont{These authors contributed equally to this work.}
\author[2]{\fnm{Haozhi} \sur{Cao}}\email{haozhi001@e.ntu.edu.sg}
\equalcont{These authors contributed equally to this work.}
\author[3]{\fnm{Jianxiong} \sur{Yin}}\email{jianxiongy@nvidia.com}
\author[1]{\fnm{Zhenghua} \sur{Chen}}\email{chen0832@e.ntu.edu.sg}
\author[1]{\fnm{Xiaoli} \sur{Li}}\email{xlli@i2r.a-star.edu.sg}
\author[1]{\fnm{Zhengguo} \sur{Li}}\email{ezgli@i2r.a-star.edu.sg}
\author[4]{\fnm{Qianwen} \sur{Xu}}\email{qianwenx@kth.se}
\author*[2]{\fnm{Jianfei} \sur{Yang}}\email{yang0478@e.ntu.edu.sg}

\affil[1]{\orgdiv{Institute for Infocomm Research}, \orgname{A*STAR}, \orgaddress{\street{1 Fusionopolis Way}, \postcode{138632}, \country{Singapore}}}

\affil[2]{\orgdiv{School of Electrical and Electronic Engineering}, \orgname{Nanyang Technological University}, \orgaddress{\street{50 Nanyang Avenue}, \postcode{639798}, \country{Singapore}}}

\affil[3]{\orgname{NVIDIA AI Tech Centre}, \orgaddress{\street{3 International Business Park Rd}, \postcode{609927}, \country{Singapore}}}

\affil[4]{\orgdiv{Department of Electric Power and Energy Systems}, \orgname{KTH Royal Institute of Technology}, \orgaddress{\street{Teknikringen 33}, \city{Stockholm}, \postcode{114 28}, \country{Sweden}}}

\abstract{While action recognition (AR) has gained large improvements with the introduction of large-scale video datasets and the development of deep neural networks, AR models robust to challenging environments in real-world scenarios are still under-explored. We focus on the task of action recognition in dark environments, which can be applied to fields such as surveillance and autonomous driving at night. Intuitively, current deep networks along with visual enhancement techniques should be able to handle AR in dark environments, however, it is observed that this is not always the case in practice. To dive deeper into exploring solutions for AR in dark environments, we launched the UG\textsuperscript{2}+ Challenge Track 2 (UG2-2) in IEEE CVPR 2021, with a goal of evaluating and advancing the robustness of AR models in dark environments. The challenge builds and expands on top of a novel ARID dataset, the first dataset for the task of dark video AR, and guides models to tackle such a task in both fully and semi-supervised manners. Baseline results utilizing current AR models and enhancement methods are reported, justifying the challenging nature of this task with substantial room for improvements. Thanks to the active participation from the research community, notable advances have been made in participants' solutions, while analysis of these solutions helped better identify possible directions to tackle the challenge of AR in dark environments.}


\maketitle

\section{Introduction}
\label{section:intro}

The emergence of various large-scale video datasets, along with the continuous development of deep neural networks have vastly promoted the development of video-based machine vision tasks, with action recognition (AR) being one of the spotlights. Recently, there have been increasing applications of automatic AR in diverse fields, e.g., security surveillance \citep{chen2010real,zou2019wifi,ullah2021efficient}, autonomous driving \citep{royer2007monocular,cao2019bypass,chen2020survey}, and smart home \citep{fahad2015integration,feng2017smart,yang2018device}. As a result, effective AR models that are robust to the different environments are required to cope with the different real-world scenarios. There has indeed been a significant improvement in the performance of AR models, reaching superior accuracies across various datasets \citep{wang2019hallucinating,ghadiyaram2019large,gowda2021smart}. 

{
\color{black}
Despite the rapid progress made by current AR research, most research aims to improve the model performance on existing AR datasets that are constrained by several factors. One of which concerns the fact that current AR datasets (e.g.\ HMDB51 \citep{kuehne2011hmdb}, UCF101 \citep{soomro2012ucf101}, and Kinetics \citep{kay2017kinetics}) are constructed with online videos which are generally shot under a non-challenging environment, with adequate illumination and contrast. The existence of such constraints could lead to the observable fragility of proposed methods, which are not capable to generalize well to adverse environments, including dark environments with low illumination, and is thus eminently related to the degrading performance of AR models in dark environments. Take security surveillance as an example: automated AR models could play a vital role in anomaly detection. However, anomaly actions are more common at night time and in dark environments, yet current AR models are obscured by darkness, and are unable to recognize any actions effectively. Autonomous systems are another example, where darkness has hampered the effectiveness of onboard cameras so severely that most vision-based autonomous driving systems are strictly prohibited at night \citep{brown2019autonomous}, while those who do allow night operation could cause severe accidents \citep{boudette2021it}.
}

To mitigate performance degradation of AR models in dark environments, one intuitive method is to perform pre-processing of dark videos which could improve the visibility of the dark videos. Such a method is indeed effective from the human vision perspective. Over the past decade, various visual enhancement techniques \citep{guo2016lime,ying2017new,zhang2019kindling,guo2020zero,li2021learning} have been proposed to improve the visibility of degraded images and videos, ranging from dehazing, de-raining to illumination enhancements. Given the effectiveness of deep neural networks in related tasks such as image reconstruction, deep-learning based illumination enhancement methods have also been developed with the introduction of various illumination enhancement datasets (e.g., SID \citep{chen2018learning}, ReNOIR \citep{anaya2018renoir} and LOL dataset \citep{liu2021benchmarking}). The results are reportedly promising from a human vision viewpoint, given their capability in improving the visual quality of low-illumination images and videos.

{
\color{black}
In spite of their capability in generating visually enhanced images and videos, prior research \citep{xu2021arid,singh2022action,chen2021darklight} has shown that a majority of illumination enhancement methods are incapable of improving AR performance in dark videos consistently. This is caused by two aspects: first, most illumination enhancement methods are developed upon low-illumination images, which are static and do not contain motion information. For the few illumination enhancement video datasets (e.g, DRV \citep{chen2019seeing}), videos collected are also mostly static, with the ``ground truth" of the dark videos shot by long exposures. Therefore, human actions are generally not included in the current video datasets for illumination enhancement.
}

Second, current illumination enhancement datasets target predominantly on human vision, with the evaluation of method based not only on quantitative evaluation but also on rather subjective qualitative evaluation (e.g., US \citep{guo2020zero} and PI scores \citep{mittal2012making,ma2017learning,blau2018perception}). Quantitative evaluation of illumination enhancement methods is also based mostly on the quality of the image/video (e.g., PSNR) instead of the understanding of image/video (e.g., classification and segmentation). The misalignment between the target of applying illumination enhancements to dark videos for AR and that of the illumination enhancement datasets would therefore be unable to guide illumination enhancement methods to improve on AR accuracies in dark videos.

{
\color{black}
To apply AR models in real-world practical applications, the model is expected to be robust to videos shot in all environments, including the challenging dark environments. In view of the inability of current solutions in addressing AR in dark environments, it is therefore highly desirable to conduct comprehensive research on effective methods to cope with such challenging environments. Such research could enable models to handle real-world dark scenarios, and benefit in various fields such as security and autonomous driving.
}

To bridge the gap between the lack of research in AR models robust to dark environments and the wide application in real-world scenarios of such research, we propose the UG\textsuperscript{2}+ Challenge Track 2 in IEEE CVPR 2021. The UG\textsuperscript{2}+ Challenge Track 2 (UG2-2) aims to evaluate and advance the robustness of AR models in poor visibility environments, focusing on dark environments.
\textcolor{black}{
Specifically, UG2-2 is structured into two sub-challenges, featuring different actions and diverse training protocols. UG2-2 is built on top of a recent AR dataset: ARID, which is a collection of realistic dark videos dedicated to AR. UG2-2 further expands the original ARID, strengthening its capability of guiding models in recognizing actions in dark environments.
}
More specific dataset details and evaluation protocols are illustrated in Section \ref{section:track2:arid}. Compare with previous works and challenges, UG2-2 and its relevant datasets include the following novelties:
\begin{itemize}
    \item \textit{Addressing Videos from Dark Environments:} The dataset utilized in UG2-2 is the first video dataset dedicated to action recognition in the dark. The original dataset with its expansion is collected from real-world scenarios. It provides much-needed resources to research actions captured in the challenging dark environments, and to design effective recognition methods robust towards dark environments.
    \item \textit{Covering Fully and Semi-Supervised Learning:} The two sub-challenges in UG2-2 are structured to cover both fully supervised learning (UG2-2.1) and semi-supervised learning (UG2-2.2). To the best of our knowledge, this is the first challenge that involves semi-supervised learning of dark videos. While our dataset provides resources for AR in dark environments, more feasible and efficient strategies to learn robust AR models is to adapt or generalize models learnt in non-challenging environments (which usually are of larger scale) to the dark environments. In this sense, our challenge promotes research into leveraging current datasets to boost performance on dark videos.
    \item \textit{Greatly Challenging:} Compare with conventional AR datasets (e.g., UCF101), the dataset utilized in the fully supervised sub-challenge is of small scale. Yet the winning solution of this sub-challenge achieves a performance inferior to that in UCF101. Meanwhile, even though the cross-domain video dataset used in the semi-supervised sub-challenge is comparable to conventional cross-domain video dataset (i.e., UCF-HMDB \citep{sultani2014human}), the winning solution performance is also inferior to that achieved in UCF-HMDB. Performances of second runner-up solutions of the semi-supervised sub-challenge are of a large gap away from the winning solution. The results prove that our datasets are greatly challenging with a large room for further improvements.
\end{itemize}

The rest of this article is organized as follows: Section \ref{section:related} reviews previous action recognition and dark visual datasets, as well as various action recognition methods. Section \ref{section:track2} introduces the details of the UG2-2 challenge, with its dataset, evaluation protocol and baseline results. Further, Section \ref{section:results} illustrates the results of the competition and related analysis, while briefly discussing the reflected insights as well as possible future developments. The article is concluded in Section \ref{section:concl}.

\section{Related Works}
\label{section:related}

\subsection{Large-Scale Datasets}
\label{subsection:largeDatasets}
Various datasets have been proposed to advance the development of video action recognition (AR). Earlier datasets (e.g. KTH \citep{schuldt2004recognizing}, Weizmann \citep{gorelick2007actions}, and IXMAS \citep{weinland2007action}) comprise a relatively small number of action classes. The videos in these datasets were recorded offline performed by several actors under limited scenarios. For example, KTH \citep{schuldt2004recognizing} includes six different action classes performed by 25 actors under 4 different scenarios. With the advancing performance of deep-learning-based methods, there has been an urging demand for larger and more complicated datasets.
{
\color{black}
To address this issue, subsequent datasets, such as HMDB51 \citep{kuehne2011hmdb} and UCF101 \citep{soomro2012ucf101}, have been proposed by collecting videos from more action classes and more diverse scenarios. Specifically, HMDB51 \citep{kuehne2011hmdb} is constructed with videos of 51 action classes collected from a variety of sources from movies to online video platforms, while UCF101 \citep{soomro2012ucf101} consists of 101 different actions collected from user-uploaded videos.
}

Both HMDB51 \citep{kuehne2011hmdb} and UCF101 \citep{soomro2012ucf101} have served as the standard benchmark of AR, while they possess insufficient data variation to train deep models, mainly because they contain multiple clips sampled from the same video. To address this issue, larger datasets with more variation have been proposed. One of the most representative examples is the famous Kinetics-400 \citep{kay2017kinetics}. The Kinetics-400 incorporates 306,245 clips from 306,245 videos (i.e. each clip is from a different video) in 400 action classes. There are at least 400 clips within each class, which guarantees more inner-class variety compared to other datasets. The following versions of Kinetics dataset, including Kinetics-600 \citep{carreira2018short} and Kinetics-700 \citep{carreira2019short}, have also been collected abiding a similar protocol. In addition to Kinetics datasets, many large-scale datasets are presented to increase the variety of samples from different perspective, such as Something-Something \citep{goyal2017something} for human-object interactions, AVA \citep{gu2018ava} for localized actions, Moments-in-Time \citep{monfort2019moments} for both visual and auditory information. While the emerging large-scale datasets push the performance limit of deep models, most of them are mainly collected from internet or shot under normal illuminations.

\subsection{Dark Visual Datasets}
\label{subsection:darkDatasets}
There have been emerging research interests towards high-level tasks in low-illumination environments in the field of computer vision. This increasing attention leads to a number of image-based datasets in dark environments. The earlier datasets were mainly designed for image enhancement or restoration, which include LOL \citep{wei2018deep}, SID \citep{chen2018learning}, ExDARK \citep{loh2019getting} and DVS-Dark \citep{zhang2020DVS}. Specifically, LOL \citep{wei2018deep} and SID \citep{chen2018learning} consist of pairs of images shot under different exposure time or ISO, while ExDARK \citep{loh2019getting} contains images collected from various online platforms. DVS-Dark consists of event images instead of RGB images, which can respond to changes in brightness, and the recent work \citep{lv2021attention} proposed to further extend the scale of the dataset by introducing synthetic low-light images. These research interests have also expanded to the video domain.
{
\color{black}
Several video datasets, such as DRV \citep{chen2019seeing} and SMOID \citep{jiang2019learning}, have been proposed specifically for low-light video enhancement, which include raw videos captured in dark environments and corresponding noise-free videos obtained by using long-exposure. However, these datasets mainly encompass static scenes with trivial dynamic motion and therefore are not suitable for AR which significantly relies on motion information~\citep{kong2022human,beddiar2020vision,li2020tea}. Furthermore, both datasets are of small scales (e.g. 179 samples for DRV \citep{chen2019seeing} and 202 samples for SMOID \citep{jiang2019learning}). In this paper, we introduce the ARID dataset \citep{xu2021arid} and its variants, which contain more samples of various actions and designed specifically for action recognition in dark videos, as our evaluation benchmark. A detailed comparison between the introduced ARID dataset with DRV and SMOID datasets is displayed in Table \ref{table:2-a-ARID_vs_prev}. We also display the sampled frames from ARID, DRV, and SMOID as in Fig. \ref{figure:2-a-samples_ARID_others}, which shows that both DRV and SMOID includes videos that mainly encompass static scenes and are not suitable for the action recognition task.

\begin{figure*}[!t]
    \color{black}
    \centering
    \includegraphics[width=.75\linewidth]{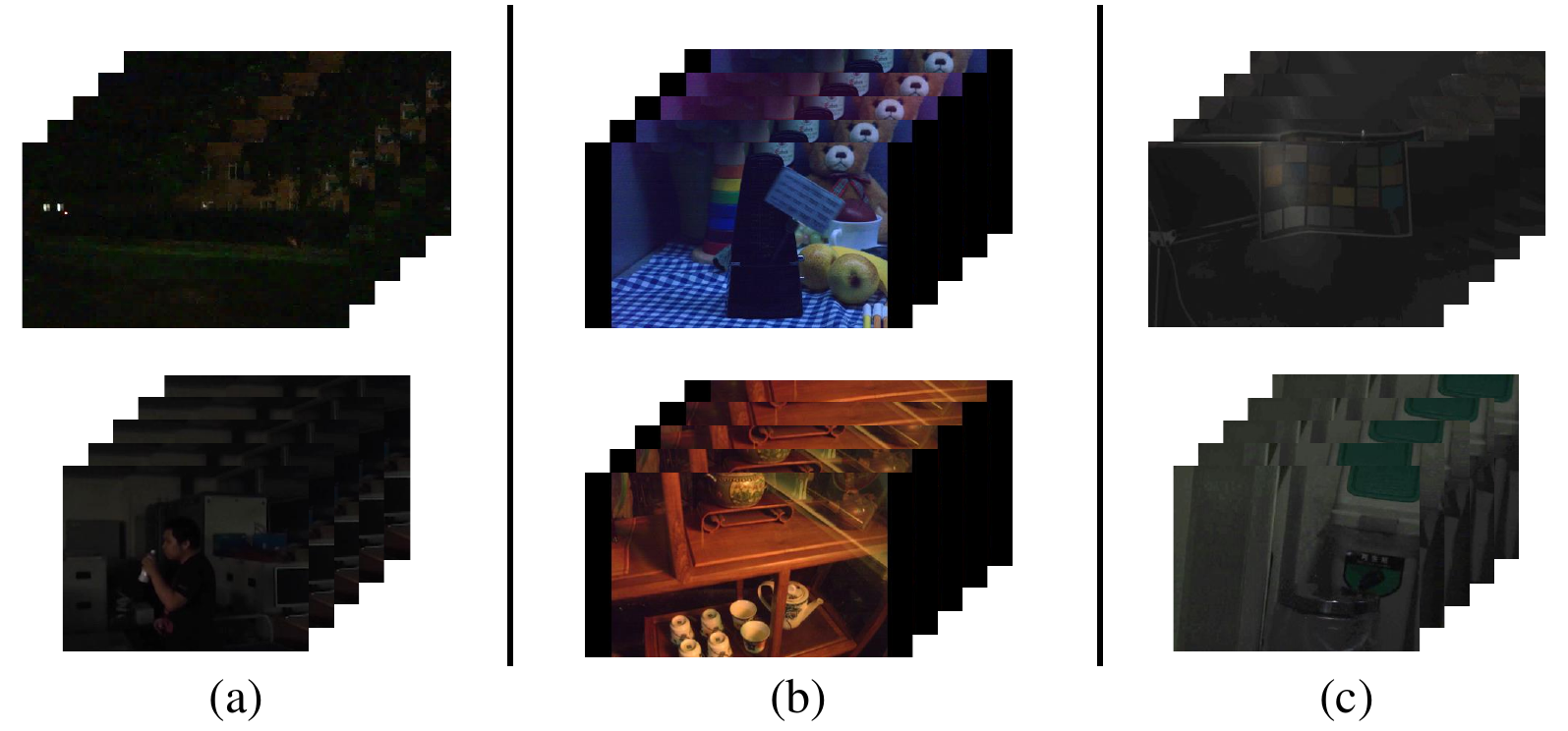}
    \caption{Sampled frames from (a) ARID, (b) DRV, and (c) SMOID. Note that all sampled frames from the dark videos have been tuned much brighter for visualization.}
\label{figure:2-a-samples_ARID_others}
\end{figure*}

\begin{table}[!t]
\color{black}
\centering
\caption{Compare ARID and prior dark video datasets.}
\resizebox{1.\linewidth}{!}{
\begin{tabular}{c|c|m{.5\textwidth}|m{.35\textwidth}|m{.3\textwidth}}  
\hline
\hline
Dataset & Publication & Task & Size & Evaluation \\
\hline
ARID
& IJCAIW-21 & Action recognition in dark videos & 3,748 video clips & Classification accuracy (\%)\tstrut\bstrut\\
\hline
DRV
& ICCV-19 & Deep processing and enhancement of extreme low-light raw videos & 202 static raw videos & Quality evaluation metrics: PSNR (dB), SSIM, MAE\tstrut\bstrut\\
\hline
SMOID
& ICCV-19 & Processing and enhancement of low-light RGB videos & 179 low-light/well-lighted video pairs (35800 images) & Quality evaluation metrics: PSNR (dB), SSIM, MABD\tstrut\bstrut\\
\hline
\hline
\end{tabular}
}
\label{table:2-a-ARID_vs_prev}
\end{table}
}

\subsection{Action Recognition Methods}
\label{subsection:ARmethods}
In the era of deep learning, early state-of-the-art AR methods are fully supervised methods mainly based on either 3D CNN \citep{ji20123d} or 2D CNN \citep{karpathy2014large}. 3D CNN \citep{ji20123d} attempts to jointly extract the spatio-temporal features by expanding the 2D convolution kernel to the temporal dimension, while this expansion suffers from high computational cost. To alleviate this side effect, subsequent works, such as P3D \citep{qiu2017learning} and R(2+1)D \citep{tran2018closer}, improve the efficiency by replacing 3D convolution kernels with pseudo 3D kernels. As for 2D CNN, due to the lack of temporal features, early 2D-based methods \citep{simonyan2014twostream} usually require additional hand-crafted features as input (e.g. optical flow) to represent the temporal information. More recent methods attempt to model the temporal information in a learnable manner. For example, TSN \citep{wang2016temporal} proposed to extract more abundant temporal information by utilizing a sparse temporal sampling strategy. SlowFast networks \citep{feichtenhofer2019slowfast} proposed to utilize dual pathways with slow or high temporal resolutions to extract spatial or temporal features, respectively. 

\textcolor{black}{
The outstanding performances of fully supervised methods mainly relies on large-scale labeled datasets, whose annotations are resource-expensive. Moreover, networks trained in the fully-supervised manner suffer from poor transferability and generalization. To increase the efficiency and generalization of extracted features, some works proposed to utilize semi-supervised approaches, such as self-supervised learning \citep{fernando2017self,xu2019self,yao2020video,wang2020self} and Unsupervised Domain Adaptation (UDA) \citep{pan2020adversarial,munro2020multi,choi2020shuffle,xu2021partial}. 
}
Self-supervised learning is designed to extract effective video representation from unlabeled data. The core of self-supervised learning is to design a pretext task to generate supervision signals through the characteristic of videos, such as frame orders \citep{fernando2017self,xu2019self} and play rates \citep{yao2020video,wang2020self}. On the other hand, UDA aims to extract the transferable representation across the labeled data in the source domain and the unlabeled data in the target domain. Compared to image-based UDA methods \citep{ganin2015unsupervised,ganin2016domain,busto2018open}, there exists fewer works in the field of video-based UDA (VUDA). \citep{pan2020adversarial} is one of the primary works focusing on VUDA, which attempts to address the temporal misalignment by introducing a co-attention module across the temporal dimension. \citep{munro2020multi} further leverages the multi-modal input of video to tackle VUDA problem. SAVA \citep{choi2020shuffle} proposed an attention mechanism to attend to the discriminate clips of videos and PATAN \citep{xu2021partial} further expanded the UDA problem to a more general partial domain adaption problem. In this work, we structured sub-challenges by covering both fully-supervised and semi-supervised to inspire novel AR methods in poor visibility environments.

{
\begin{table}[!t]
\color{black}
\centering
\caption{Compare ARID and ARID-\textit{plus}.}
\resizebox{.7\linewidth}{!}{
\begin{tabular}{c|m{.22\textwidth}|m{.4\textwidth}}  
\hline
\hline
Comparison & ARID & ARID-\textit{plus} \\
\hline
Task
& Action recognition & Action recognition \tstrut\bstrut\\
\hline
Supervision
& Fully & Fully (UG2-2.1) and semi (UG2-2.2) \tstrut\bstrut\\
\hline
\# of Videos
& 3,784 & 3,226 (UG2-2.1); 2,335 (UG2-2.2) \tstrut\bstrut\\
\hline
\# of Classes
& 11 & 6 (UG2-2.1); 5 (UG2-2.2) \tstrut\bstrut\\
\hline
Resolution
& 320$\times$240 & 320$\times$240 and 426$\times$240 \tstrut\bstrut\\
\hline
\hline
\end{tabular}
}
\label{table:3-a-ARID_vs_ARID_plus}
\end{table}
}
{
\begin{figure}[t]
\color{black}
    \centering
    \includegraphics[width=1.\linewidth]{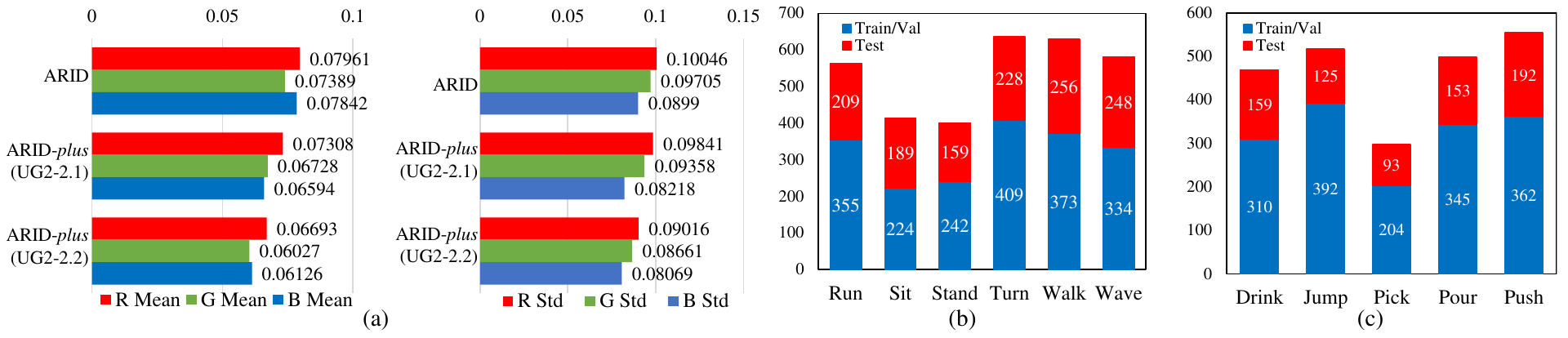}
    \caption{(a) Bar charts of the RGB mean (left) and standard deviation (right) values for ARID and the expanded dataset for UG2-2: ARID-\textit{plus}. The statistics for the two sub-challenges UG2-2.1 and UG2-2.2 are separated. All values are normalized to the range of [0.0. 1.0]. (b) The distribution of clips among action classes in ARID-\textit{plus} (UG2-2.1). (c) The distribution of clips among action classes in ARID-\textit{plus} (UG2-2.2). For (b) and (c), the blue and red bars indicate the number of clips in the training/validation and testing partitions. Best viewed in color.}
\label{figure:3-1-distribution_mean_std}
\end{figure}
}

{
\begin{table}[!t]
\color{black}
\centering
\caption{Results of AR models on the original ARID dataset.}
\resizebox{.8\linewidth}{!}{
\begin{tabular}{c|c|cc}  
\hline
\hline
Models    & Overall & \textit{Singular Person Actions} & \textit{Actions with Objects} \\
\hline
I3D-RGB         & 60.27\% & 44.78\% & 65.88\%\tstrut \\
3D-ResNet-50    & 70.84\% & 66.56\% & 74.71\% \\
3D-ResNeXt-101  & 74.73\% & 70.22\% & 78.81\% \\
SlowOnly        & 75.70\% & 71.01\% & 80.29\%\bstrut \\
\hline
\hline
\end{tabular}
}
\label{table:3-b-ARID_ori_results}
\end{table}
}

\section{Introduction of UG\texorpdfstring{\textsuperscript{2}}{\texttwoinferior}+ Challenge Track 2}
\label{section:track2}

The UG\textsuperscript{2}+ Challenge Track 2 (UG2-2) aims to evaluate and advance the robustness of AR methods in dark environments. In this section, we detail the datasets and evaluation protocols used in UG2-2, as well as the baseline results for either sub-challenges. The datasets of UG2-2 for either sub-challenges are built based on the Action Recognition In the Dark (ARID) dataset. We begin this section by a brief review of the ARID dataset.

\subsection{The ARID Dataset}
\label{section:track2:arid}
The ARID dataset \citep{xu2021arid} is the first video dataset dedicated to action recognition in dark environments. The dataset is a collection of videos shot by commercial cameras in dark environments, with actions performed by 11 volunteers. In total, it comprises 11 action classes, including both \textit{Singular Person Actions} (i.e., jumping, running, turning, walking, and waving) as well as \textit{Actions with Objects} (i.e., drinking, picking, pouring, pushing, sitting, and standing). The dark videos are shot in both indoor and outdoor scenes with varied lighting conditions. The dataset consists of a total of 3,784 video clips, with the minimum action class containing 205 video clips. The clips of every action class are divided into clip groups according to the different actors and scenes. Similar to previous action recognition datasets (e.g., HMDB51 \citep{kuehne2011hmdb} and UCF101 \citep{soomro2012ucf101}), three train/test splits are selected, with each split partitioned according to the clip groups, with a ratio of $7:3$. The splits are selected to maximize the possibility that each clip group is presented in either the training or testing partition. All video clips in ARID are fixed to a 30 FPS frame rate, and a unified resolution of $320\times240$. The overall duration of all video clips combined is 8,721 seconds.

{
\color{black}
To gain a more comprehensive understanding over the challenges posed by recognizing actions in the ARID dataset, we examine the performance of four AR models on the ARID, namely I3D \citep{carreira2017quo}, 3D-ResNet-50 \citep{hara2018can}, 3D-ResNeXt-101 \citep{hara2018can}, and SlowOnly \citep{feichtenhofer2019slowfast}. We report both the overall top-1 results as well as the top-1 results of each subset of actions (\textit{Singular Person Actions} and \textit{Actions with Objects}), as displayed in Table \ref{table:3-b-ARID_ori_results}. The results show that while various models could reach an accuracy of over 70\%, there is a notable gap between the accuracy of action classes with objects and that of singular person actions. While singular person actions such as ``running'' and ``walking'' require temporal-based reasoning for classification, actions with objects could be classified via observing particular objects (e.g., a bottle or cup for the action ``drinking''.) The gap is therefore more significant for models with poorer temporal-reasoning ability such as I3D.
}

Though the ARID dataset pioneers the investigation of action recognition methods in dark environment, it has its own limitations. Compared with current SOTA benchmarks such as Kinetics \citep{kay2017kinetics} and Moments-in-Time \citep{monfort2019moments}, the ARID is of limited scale, especially in terms of the number of videos per class. The limited scale of ARID prohibits complex deep learning methods to be trained, owing to a higher risk of overfitting. Increasing the dataset scale is an effective solution for such constraint, given that conventional action recognition dataset follows the same development path. However, the collection and annotation of dark videos is of high cost, given that there is limited public dark video on any public video platforms. Therefore, the strategy of increasing the dataset scale could only bring limited improvement to the dataset.
{
\color{black}
Given the vast availability of videos shot in non-challenging environments, such videos should be fully utilized to train transferable models that could generalize to dark videos. To this end, we introduce a comprehensive extension of the ARID dataset: ARID-\textit{plus}, to address the issues of the original ARID dataset, and serve as the datasets for the two sub-challenges of UG2-2. Table \ref{table:3-a-ARID_vs_ARID_plus} depicts a brief comparison over ARID and ARID-plus.
}

\begin{figure*}[t]
    \centering
    \includegraphics[width=1.\linewidth]{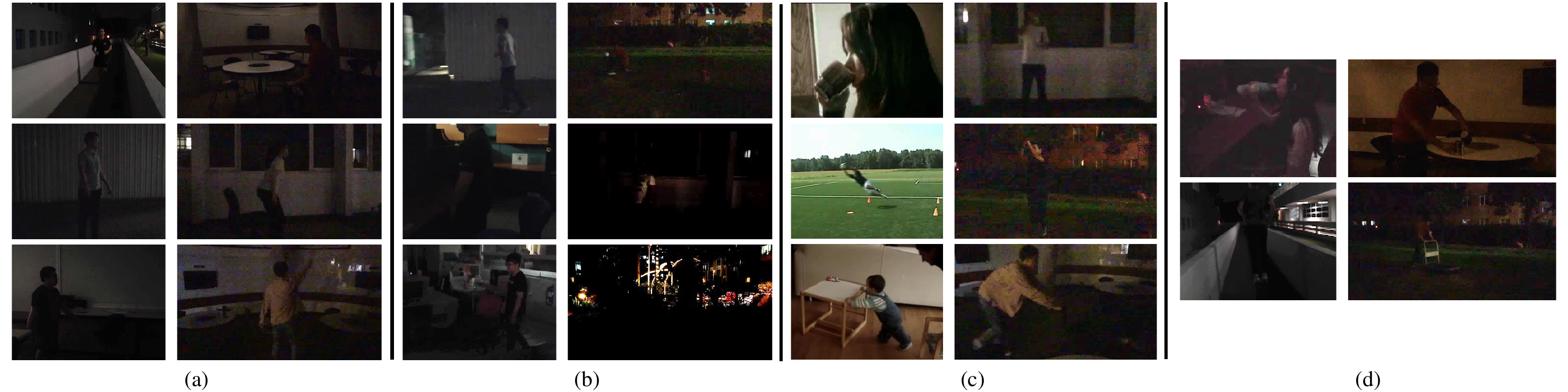}
    \caption{Sampled frames from (a) the train/validation set of ARID-\textit{plus} (UG2-2.1), (b) the hold-out test set of ARID-\textit{plus} (UG2-2.1), (c) the train/validation set of ARID-\textit{plus} (UG2-2.2) and (d) the hold-out test set of ARID-\textit{plus} (UG2-2.2). Note that all sampled frames from dark videos have been tuned much brighter for visualization.}
\label{figure:3-2-samples}
\end{figure*}

{
\begin{table}[!t]
\color{black}
\centering
\caption{Baseline Results of AR models without enhancements for UG2-2.1.}
\resizebox{.55\linewidth}{!}{
\begin{tabular}{c|c|cc}  
\hline
\hline
Input & Models    & Top-1             & Top-5 \\
\hline
\multirow{6}{*}{RGB}
& I3D-RGB         & 21.64\%           & 85.42\%\tstrut \\
& 3D-ResNet-50    & 34.14\%           & 94.49\% \\
& 3D-ResNeXt-101  & 34.45\%           & 96.82\% \\
& TSM             & 26.37\%           & 87.82\% \\
& SlowOnly        & 27.08\%           & 91.00\% \\
& X3D-M           & 20.25\%           & 87.98\%\bstrut \\
\hline
\multirow{6}{*}{\parbox{0.12\linewidth}{\centering Optical Flow}}
& I3D-OF-TVL1          & 20.56\%           & 85.18\%\tstrut \\
& SlowOnly-OF-TVL1     & 57.25\%           & 96.82\% \\
& I3D-OF-FF          & 21.88\%           & 86.04\% \\
& SlowOnly-OF-FF     & 58.88\%           & 96.97\% \\
& I3D-OF-GMF          & 22.73\%           & 88.29\% \\
& SlowOnly-OF-GMF     & 59.74\%           & 97.13\%\bstrut \\
\hline
\multirow{6}{*}{\parbox{0.12\linewidth}{\centering Two-stream}}
& I3D-TS-TVL1          & 21.41\%           & 85.26\%\tstrut \\
& SlowOnly-TS-TVL1     & 50.61\%           & 96.74\% \\
& I3D-TS-FF          & 22.96\%           & 86.50\% \\
& SlowOnly-TS-FF     & 55.93\%           & 96.43\% \\
& I3D-TS-GMF          & 23.89\%           & 88.75\% \\
& SlowOnly-TS-GMF     & 57.02\%           & 96.66\%\bstrut \\
\hline
\hline
\end{tabular}
}
\label{table:3-1-UG2_2.1_baseline}
\end{table}
}

{
\begin{table*}[!t]
\color{black}
\centering
\caption{Baseline Results of AR models with off-the-shelf enhancements for UG2-2.1.}
\resizebox{.9\linewidth}{!}{
\begin{tabular}{l|cccccc}
\hline
\hline
Enhancements     & I3D-RGB & 3D-ResNet-50 & 3D-ResNeXt-101 & TSM & SlowOnly & X3D-M \\
\hline
GIC              & 19.08\% & 39.48\% & 50.50\% & 26.84\% & 37.47\% & 18.00\% \\
LIME             & 18.39\% & 39.02\% & 33.75\% & 27.54\% & 27.54\% & 18.00\% \\
Zero-DCE         & 18.39\% & 48.80\% & 60.28\% & 23.20\% & 27.60\% & 19.08\% \\
StableLLVE       & 19.86\% & 46.24\% & 34.45\% & 29.64\% & 30.41\% & 17.84\% \\
\hline
None             & 21.64\% & 26.37\% & 34.14\% & 34.45\% & 27.08\% & 20.25\% \\
\hline
\hline
\end{tabular}
}
\label{table:3-2-UG2_2.1_enhancement}
\end{table*}
}

\subsection{Fully Supervised Action Recognition in the Dark}
\label{section:track2:2.1}
To equip AR models the ability to cope with dark environments for applications such as night surveillance, the most intuitive method would be no other than training action models in a fully supervised manner with videos shot in the dark, which motivates the construction of Sub-Challenge 1. The first component of ARID-\textit{plus} serves as the dataset of Sub-Challenge 1 of UG2-2 (UG2-2.1), where participants are given the annotated dark videos for fully supervised action recognition. A total of 1,937 real-world dark video clips capturing actions by volunteers are adopted as the training and/or validation sets, with the recommended train/validation split provided to participants. The video clips contain six categories of actions, i.e., run, sit, stand, turn, walk, and wave. For testing, a hold-out set with 1,289 real-world dark video clips are provided, collected with similar methods as the training/validation video clips, with the same classes. In total, there are a minimum of 456 clips for each action.
{
\color{black}
A detailed distribution of train(validation)/test video clips is shown in Fig. \ref{figure:3-1-distribution_mean_std}(b). It is noted that the current dataset is non-uniform as depicted in Fig. \ref{figure:3-1-distribution_mean_std}(b), especially in terms of the number of training and validation videos. This is thanks to the varied difficulty in collecting the different categories of actions, which may cause additional biases during training.
}

During training, participants can optionally use pre-trained models (e.g., models pretrained on ImageNet \citep{deng2009imagenet} or Kinetics), and/or external data, including self-synthesized or self-collected data. If any pre-trained model or external data is used, participants must state explicitly in their submissions. The participants are ranked by the top-1 accuracy of the hold-out test set, while all the solutions of candidate winners are tested for their reproducibility.

The video clips adopted for training and testing in UG2-2.1 include that in the original ARID dataset, as well as new video clips. Several changes are adopted during the collection of the new video clips. Firstly, the new video clips are shot in a number of new scenes, whose visibility is even lower. This is justified statistically by lower RGB mean values and standard deviation (std) values as depicted in Fig. \ref{figure:3-1-distribution_mean_std}(a). Secondly, videos collected in the original ARID dataset follow a $4:3$ aspect ratio, which matches standard 320p or 480p videos. Meanwhile, the currently more common High-Definition (HD) videos would have a $16:9$ aspect ratio, with larger view angles. Following the aspect ratio of HD videos, the new video clips are fixed to a resolution of $426\times240$. We have also extended the length of each video clip, from an average 2.3 seconds per clip for clips in the original ARID to an average of 4 seconds per clip for the new video clips. Sampled frames from the train/validation set and the hold-out test set are displayed in Fig. \ref{figure:3-2-samples}(a) and Fig. \ref{figure:3-2-samples}(b).

\begin{figure}[t]
    \centering
    \includegraphics[width=.7\linewidth]{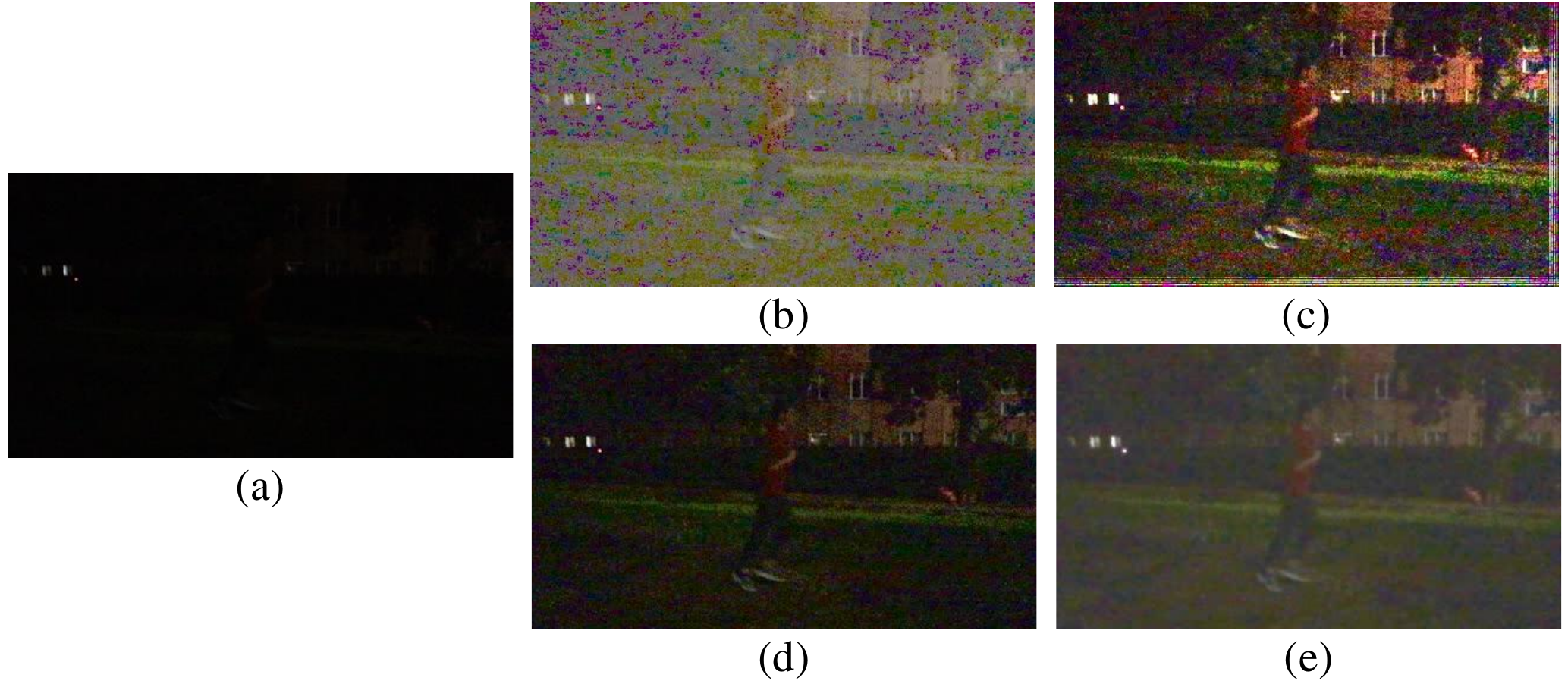}
    \caption{Comparison of (a) a sampled frame from ARID-\textit{plus} with the results after applying enhancements: (b) GIC, (c) LIME, (d) Zero-DCE, and (e) StableLLVE.}
\label{figure:3-8-2_1_enhancement}
\end{figure}

\subsection{Semi-Supervised Action Recognition in the Dark}
\label{section:track2:2.2}
While fully supervised training in dark videos allow models to cope with dark environments directly, publicly available dark videos are scarce compared with the vast amount of normal illuminated videos, which could be obtained with ease. Due to the high cost of both video collection and annotation, simply increasing the scale of dark video datasets for improving the effectiveness of fully supervised learning would not be a feasible strategy. Alternatively, the large amount of normal illuminated videos presented in previous public datasets should be utilized to train transferable models that could be generalized to dark videos. Such transfer may be further boosted with certain frame enhancements. The above strategy could be regarded as a semi-supervised learning strategy for action recognition in dark videos, which motivates the design of Sub-Challenge 2. The Sub-Challenge 2 of UG2-2 (UG2-2.2) is designed to guide participants to tackle action recognition in dark environments in a semi-supervised manner, achieved by generalizing models learnt in non-challenging environments to the challenging dark environments. 

\textcolor{black}{
To this end, the participants are provided with a subset of the labeled HMDB51 \citep{kuehne2011hmdb} that includes 643 videos from five action classes (i.e., drink, jump, pick, pour, and push), for the training of models in non-challenging environments.
}
Meanwhile, to facilitate the transfer of models, the second component of ARID-\textit{plus}, with a total of 1,613 dark video clips, is provided to the participants in an unlabeled manner, which can be optionally used at the participants' discretion for training and validation. The 1,613 clips contain the same five categories of actions. Similar to UG2-2.1, a hold-out set containing 722 real-world dark video clips with the same classes is provided for testing.
{
\color{black}
Overall, there are at least 297 clips for each action class. The detailed distribution of train(validation)/test dark video clips is shown in Fig. \ref{figure:3-1-distribution_mean_std}(c). It is also noted that the current dataset for UG2-2.2 is non-uniform as depicted in Fig. \ref{figure:3-1-distribution_mean_std}(c) thanks to the varied difficulty in collecting the different categories of actions, and may cause additional biases during training.
}

During training, participants can also optionally use pre-trained models, and/or external data, including self-synthesized or self-collected data. However, the 1,613 dark video clips provided during the training/validation phase are not allowed to be manually labeled for training (i.e., they must remain to be unlabeled). Participants are to state explicitly if any pre-trained model or external data is used, and are ranked by the top-1 accuracy of the hold-out test set with reproducibility subject to testing if the relevant solution's testing accuracy stands out. Changes in extra data that have been applied in UG2-2.1 have also been employed in the extra dark video clips for UG2-2.2. Such changes result in a similar degradation of clip visibility (as depicted in Fig. \ref{figure:3-1-distribution_mean_std}(a)), and an increase in view angles and average clip length. We show the sampled frames from the labeled HMDB51 train set, the unlabeled dark train/validation set, as well as the hold-out test set in Fig. \ref{figure:3-2-samples}(c) and Fig. \ref{figure:3-2-samples}(d).

{
\begin{table}[!t]
\color{black}
\centering
\caption{Baseline Results of AR models with domain adaptation methods for UG2-2.2.}
\resizebox{.65\linewidth}{!}{
\begin{tabular}{m{0.25\linewidth}|ccc}
\hline
\hline
Domain Adaptation     & I3D     & 3D-ResNet-50  & 3D-ResNeXt-101\tstrut\bstrut \\
\hline
Source-only           & 26.38\% & 24.10\%       & 27.42\%\tstrut\bstrut \\
\hline
DANN                  & 35.18\% & 36.14\%       & 44.04\%\tstrut \\
MK-MMD                & 35.32\% & 26.90\%       & 32.55\% \\
MCD                   & 27.15\% & 29.36\%       & 28.25\%\bstrut \\
\hline
Target-only           & 46.25\% & 75.07\%       & 76.87\%\tstrut\bstrut \\
\hline
\hline
\end{tabular}
}
\label{table:3-3-UG2_2.2_baseline}
\end{table}
}

\subsection{Baseline Results and Analysis}
\label{section:track2:baseline}
For both sub-challenges, we report baseline results utilizing off-the-shelf enhancement methods with fine-tuning of several popular pre-trained action recognition models and domain adaptation methods. It should be noted that these enhancement methods, pre-trained models and domain adaptation methods are not designed specifically for dark videos, hence they are by no means very competitive, and performance boosts are expected from participants.

\subsubsection{Fully Supervised UG2-2.1 Baseline Results}
\label{section:track2:baseline:2.1}
{
\color{black}
For UG2-2.1, we report baseline results from a total of six AR models including: I3D \citep{carreira2017quo}, 3D-ResNet-50 \citep{hara2018can}, 3D-ResNeXt-101 \citep{hara2018can}, TSM \citep{lin2019tsm}, SlowOnly \citep{feichtenhofer2019slowfast}, and X3D-M \citep{feichtenhofer2020x3d}. Among which, RGB frames are utilized as the input for all methods, while we also report the results utilizing optical flow obtained through methods including the more common TV-L1 \citep{zach2007duality}, and the more recent FlowFormer \citep{huang2022flowformer} and GMFlow \citep{xu2022gmflow}, which are applied to the I3D and SlowOnly methods. The results by class score fusion \citep{simonyan2014twostream} with both RGB frames and the different optical flow extracted are also reported.
}

Meanwhile, applying enhancement methods which improve the visibility of dark videos is an intuitive method to improve AR accuracies. Therefore, we also evaluate the above methods using RGB input with four enhancement methods: Gamma Intensity Correction (GIC), LIME \citep{guo2016lime}, Zero-DCE \citep{guo2020zero} and StableLLVE \citep{zhang2021learning}.
{
\color{black}
Specifically, GIC is a simple enhancement formulated as a power function $O = I ^ {(1/\gamma)}$, where the output and input pixel values $O$ and $I$ are normalized to a scale of $[0, 1]$ and the $\gamma$ value is set to be larger than 1. LIME enhances low-light videos by estimating and refining an illumination map for each color channel. Meanwhile, Zero-DCE estimates pixel-wise and high-order curves for dynamic range adjustment of dark videos, while StableLLVE leverages optical flow prior to indicate potential motion from single image such that the temporal consistency could be modelled for low light video enhancement.
}

{
\color{black}
All AR models, optical flow extraction methods, and enhancement methods adopt the officially released versions when applicable, where all learning-based methods are written with the PyTorch \citep{paszke2019pytorch} framework. All AR models are fine-tuned from their models pre-trained on Kinetics-400 \citep{kay2017kinetics}, and trained for a total of 30 epochs. Meanwhile, the optical extraction models FlowFormer and GMFlow are pre-trained on the Sintel dataset \citep{butler2012naturalistic}, the weights are frozen during AR model training and testing.
}
Due to the constraints in computation power, the batch size is unified for all models and set to 8 per GPU. All experiments are conducted with two NVIDIA RTX 2080Ti GPUs. The reported results are an average of five experiments. The detailed results are found in Table \ref{table:3-1-UG2_2.1_baseline} and Table \ref{table:3-2-UG2_2.1_enhancement}.

Overall, with the training settings as introduced above, current AR models performs poorly without any enhancements in UG2-2.1. 
{
\color{black}
The best performance is achieved by using optical flow input with the SlowOnly model, i.e., an accuracy of $57.25\%$ with TV-L1 optical flow and an accuracy of $59.74\%$ with optical flow obtained through GMFlow. In comparison, the evaluated models could achieve at least $70\%$ accuracy on the large-scale Kinetics dataset, and over $80\%$ accuracy on the HMDB51 dataset. It is worth noting that newer models (e.g., X3D-M) which produce SOTA results on large-scale datasets may perform inferior to previous models (e.g., 3D-ResNeXt-101). Therefore novel AR models may not be more generalizable than prior AR models. It is further observed that extracting more effective optical flow could bring noticeable improvements on model performance. Optical flow depicts the apparent motion of objects, which corresponds to the temporal information of human action in the case of videos in ARID. Compared to TV-L1, GMFlow estimates optical flow through a global matching formulation which address both occluded and out-of-boundary pixels more effectively. With more effective optical flow, better temporal information is obtained through the models, thus resulting in higher classification accuracy.
}

Meanwhile, the results after applying enhancements show that the evaluated enhancements may not bring consistent improvements in action recognition accuracy. The evaluated enhancements all produce visually clearer videos, where actions are more recognizable by humans, as shown in Fig. \ref{figure:3-8-2_1_enhancement}. The actor who is running can be seen visually in all sampled frames with enhancements, while the actor is almost unrecognizable in the original dark video. However, at least three AR models produce inferior performance when applying any enhancement. The best result is obtained with 3D-ResNeXt-101 while applying Zero-DCE enhancement. In general, Zero-DCE results in the best average improvement of $5.57\%$. Meanwhile, the susceptibility of each model varies greatly. 3D-ResNet-50 gains the most positive effect of $17.02\%$ average accuracy gain with enhancements applied, while TSM is most susceptible to negative effects with an average loss of $7.65\%$ accuracy. 

We argue that the negative effect of applying enhancements results from the noise brought by enhancements. Though enhanced videos are clearer from human perspectives, some enhancements break the original data distribution, and can therefore be regarded as artifacts or adversarial attacks for videos. The change in data distribution and the addition of noise could result in a notable decrease in performance for AR models. The deficiencies of the examined enhancements suggest that simple integration of frame enhancements may not be sufficient. Instead, other techniques such as domain adaptation or self-supervision could be further employed to improve the effectiveness of frame enhancements.

{
\color{black}
In short, the above observations suggest that for the fully supervised UG2-2.1, the best results are obtained with an adequate backbone (e.g., SlowOnly or 3D-ResNeXt-101) while leveraging effective optical flow as input. The intuitive approach for improving AR accuracy by applying enhancement method does not apply to any model-enhancement pairs. While Zero-DCE does show the best average improvements, it would also incur negative affect when applying to 3 out of the 6 AR models examined.
}

\subsubsection{Semi-Supervised UG2-2.2 Baseline Results}
\label{section:track2:baseline:2.2}
For UG2-2.2, we report baseline results with three AR models: I3D, 3D-ResNet-50, and 3D-ResNeXt-101. To transfer networks from the labeled normal videos to unlabeled dark videos, we employ and evaluate three different domain adaptation methods: the adversarial-based DANN \citep{ganin2015unsupervised}, and the discrepancy-based MK-MMD \citep{long2015learning} and MCD \citep{saito2018maximum}. We also examine both the source-only scenario (i.e., fully supervised learning) and target-only scenario (i.e., without any domain adaptation method). Similar to the baseline experiments in UG2-2.1, all models are pre-trained on Kinetics-400, with the whole training process set to 30 epochs. For all AR models, we freeze the first three convolutional layers, and the batch size is set to 8 per GPU. The experiments are conducted with the same hardware and framework as that of UG2-2.1 baselines. No enhancement method is employed when conducting the baseline experiments for UG2-2.2. The reported results are an average of five experiments. Detailed results are shown in Table \ref{table:3-3-UG2_2.2_baseline}.

{
\color{black}
The results in Table \ref{table:3-3-UG2_2.2_baseline} imply that though all three adaptation methods can improve the generability of the respective AR models, scoring higher than the source-only scenarios, all have a large gap towards the target-only accuracies, which are the upper bounds of the networks' performances. The large performance gap towards the upper bound also justifies the fact that there exists a large domain gap between videos shot in non-challenging environments and videos shot in dark environments. Among the three adaptation methods, DANN produces the best performance in general, resulting in an average performance gain of $12.82\%$ towards the models' source only performances.
}
The best baseline result is obtained with 3D-ResNeXt while applying DANN as the domain adaptation method. It should be noted that no enhancements or other training tricks are applied when obtaining the baseline results for UG2-2.2. Therefore, it is expected that participants could score higher than the target-only accuracies in Table \ref{table:3-3-UG2_2.2_baseline}.

\section{Results and Analysis}
\label{section:results}

\begin{figure}
    \centering
    \subfloat[One-stream Structure]{
    \includegraphics[width=0.8\linewidth]{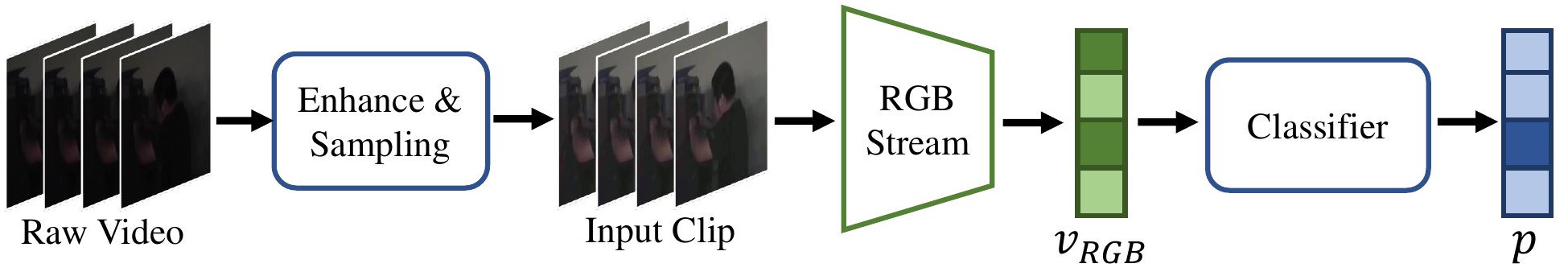}
    \label{sub-figure:4-9-1}}\\
    \subfloat[Two-stream Structure i]{
    \includegraphics[width=0.8\linewidth]{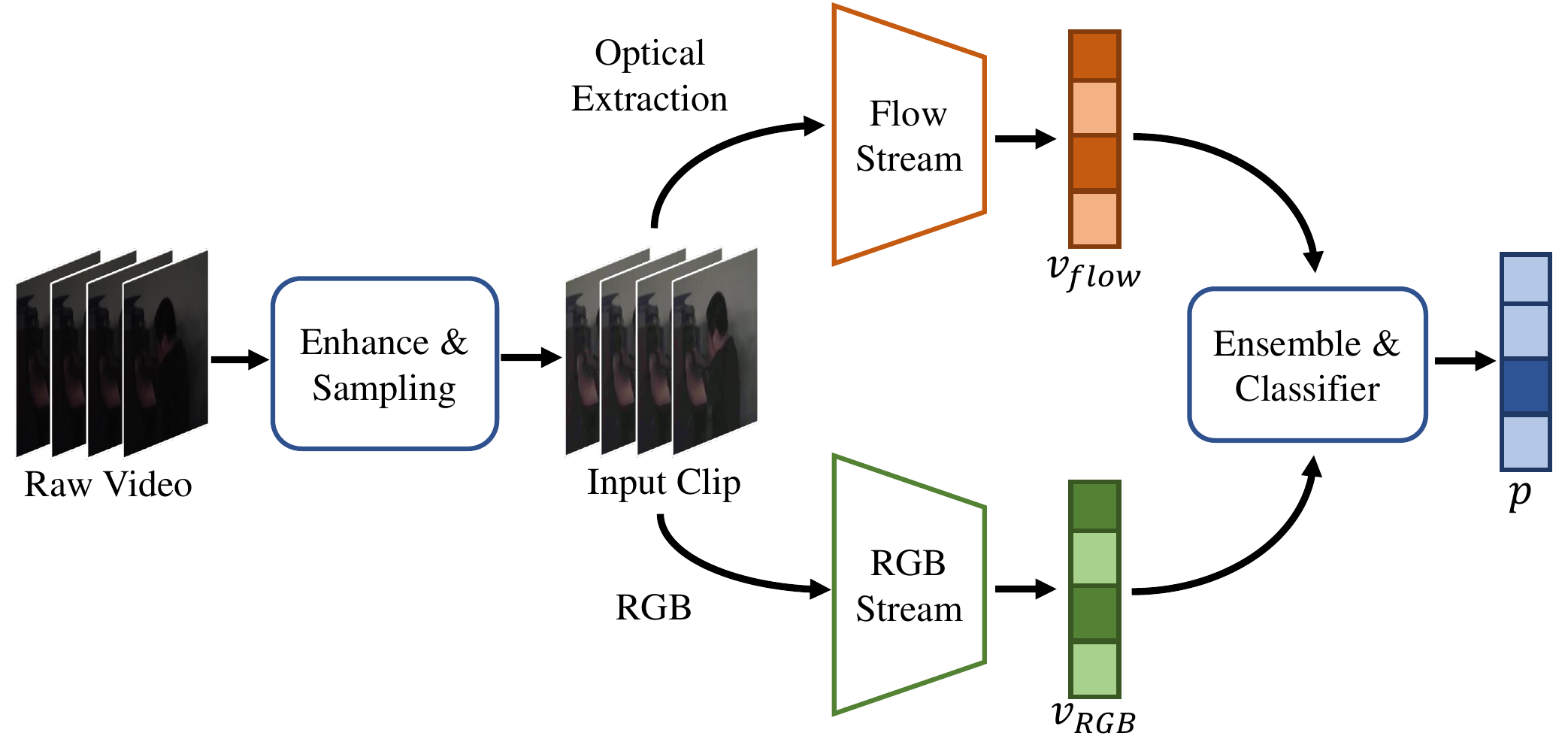}
    \label{sub-figure:4-9-2}}\\
    \subfloat[Two-stream Structure ii]{
    \includegraphics[width=0.7\linewidth]{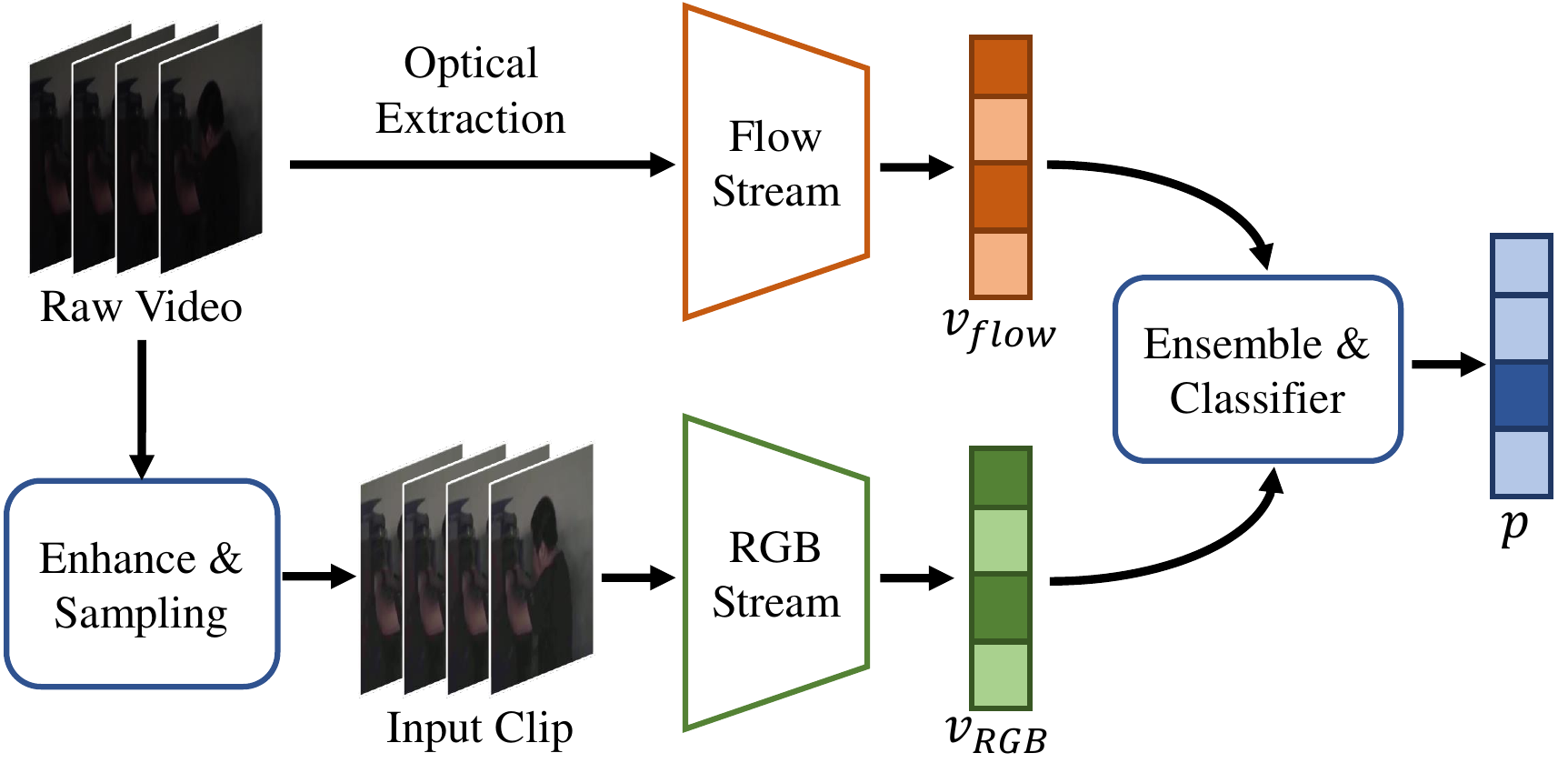}
    \label{sub-figure:4-9-3}}
    \caption{Structures utilized in UG2-2.1. Here $v_{RGB}$, $v_{flow}$ are vector features extracted from RGB and flow input, respectively. $p$ is the prediction of the final classifier.}
    \label{figure:4-9-2_1_structure}
\end{figure}

A total of 34 teams registered in the UG\textsuperscript{2}+ Challenge Track 2 (UG2-2) at CVPR, among which 25 and 12 teams submitted their results to the fully supervised sub-challenge (UG2-2.1) and the semi-supervised sub-challenge (UG2-2.2), respectively. For each sub-challenge, the team with the highest performance is selected as the winner. In this section, we summarize the technical details of some outstanding performers and compare them with our baseline results. The full leaderboards can be found in the website\footnote{\url{https://cvpr2021.ug2challenge.org/leaderboard21_t2.html}}.

\subsection{UG2-2.1: Fully Supervised Learning}
\label{section:results:2.1}
Among the 25 teams that successfully participated in this sub-challenge, 11 teams proposed novel models that outperform our baseline results. Among them, 7 teams are included in our leaderboard, where the winner team \textit{AstarTrek} achieved the best performance of $93.72\%$. While all teams constructed their models based on complex backbones, some interesting observations are as follows: (i) besides RGB, 3 out of 6 teams in the leaderboard utilized additional optical flow as input, while this extra modality did not bring solid improvement compared to those using pure RGB input; (ii) teams achieving top performance utilized low-light enhancement methods; (iii) except for the winner team, all teams trained their model from scratch with large epoch numbers (more than 200) rather than utilizing other pre-trained models, surpassing our baseline results by at least $18.83\%$.

The winner team \textit{AstarTrek} adopted a two-stream structure as shown in Fig. \ref{sub-figure:4-9-2}. 
\textcolor{black}{
The team first utilized the Gamma Intensity Correction (GIC) with $\gamma=3$ to enhance the illumination level of videos. Subsequently, both RGB and optical flow were generated as the input of the two-stream structure. Specifically, the SlowFast Network \citep{feichtenhofer2019slowfast} (based on ResNet-50 \citep{he2016deep}) pretrained on Kinetics-400 (K400) was adopted as the backbone for the RGB stream to extract spatial features from raw RGB input. For the flow stream, the team leveraged a ResNet-50-based I3D \citep{carreira2017quo} pretrained on K400 to extract temporal information from optical flow.
}
During the training process, the team adopted a two-stage procedure, where each stream was trained independently to ensure that each of them can provide reliable predictions by itself. Each stream was trained with stochastic gradient descent (SGD) with a momentum of 0.9 and a weight decay of 0.0001. The batch size was set to 32 and the initial learning rate was 0.001, decayed by a factor of 0.1 at epochs 60 and 100 (with total epochs of 800). Each input (RGB or optical flow) was first resized to a square of the height randomly sampled from [224, 288], then randomly cropped into a square of size 224$\times$224, followed by a horizontal flip with a probability of 0.5. During inference, each input (RGB or optical flow) was resized to the size of 240$\times$300. The final prediction is the average of results from both streams.

On the other hand, the runner-up team \textit{Artificially Inspired} adopted different backbones and strategies, achieving a competitive performance of $92.32\%$. As shown in Fig. \ref{sub-figure:4-9-1}, taking pure RGB as input, the team utilized Zero-DCE \citep{guo2020zero} as their enhancement method and R(2+1)D-Bert \citep{kalfaoglu2020late} as their single stream backbone. In fact, they are the only participant in the leaderboards that utilized deep-model-based method to improve the quality of dark videos. Moreover, noticing samples in ARID containing a relatively small number of frames, the team utilized Delta Sampling strategy that constructed the input sample by various sample rates while avoiding loop sampling. The team utilized 4,500 different images to train the Zero-DCE model, where 2,500 images were randomly sample from ARID dataset and the others are of different illumination levels collected from other datasets. During the training process, videos were first enhanced by the frozen Zero-DCE model to enhance their light levels and then resized to 112$\times$112. The team also included a random horizontal flip and rotation to increase the variation of input samples. According to their ablation studies, the utilization of Zero-DCE can bring an improvement of $2.98\%$ and the proposed sampling strategy surpassed other alternatives. More technical details can be refer to their report \citep{hira2021delta}.

\textcolor{black}{
Besides \textit{AstarTrek}, there are two other teams in the leaderboard, \textit{Cogvengers} and \textit{MKZ5}, which attempted to leverage the optical flow as the additional input to improve the performance. However, their performance is surpassed by most of teams taking pure RGB input by more than $2.5\%$, mainly because they utilized inferior strategies for extracting and processing optical flow.
}
Specifically, \textit{MKZ5} utilized a different two-stream structure as shown in Fig. \ref{sub-figure:4-9-3}, which directly extracted optical flow from dark videos, while \textit{AstarTrek} extracted optical flow from enhanced videos. The direct extraction may end up a worse quality of optical flow since most of the optical flow estimation methods show poor performance with low-light data \citep{zheng2020optical}. As for \textit{Cogvengers}, while they adopted the structure in Fig. \ref{sub-figure:4-9-2} similar to the winner team, they follow a one-stage training strategy to jointly optimized the two-stream model, which might be the reason for their performance gap compared to others. In addition to the two-stream models based on RGB and optical flow, the team \textit{White give} proposed another interesting two-stream structure based on pure RGB input \citep{chen2021darklight}. Specifically, adopting a similar structure in Fig. \ref{sub-figure:4-9-3}, the team replaced the flow stream with a shared-weight RGB stream taking original dark clips as input. The features from enhanced clips and dark clips were then ensembled by a self-attention module \citep{wang2018non} to extract the effective spatio-temporal features from dual streams.

\subsection{UG2-2.2: Semi-Supervised Learning}
\label{section:results:2.2}
\begin{figure}
    \centering
    \subfloat[Pseudo-label Based Structure]{
    \includegraphics[width=0.97\linewidth]{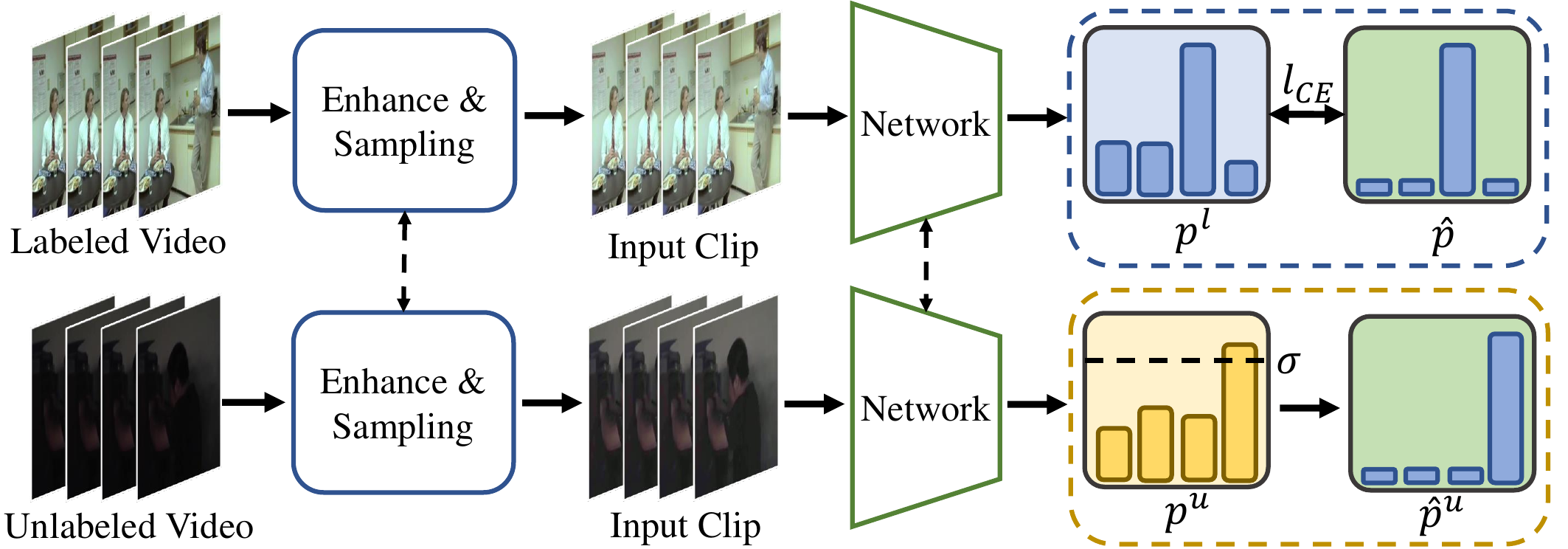}
    \label{sub-figure:4-10-1}}\\
    \subfloat[Structure of the TCL-based method. Details can be found in \citep{singh2021semi}.]{
    \includegraphics[width=0.97\linewidth]{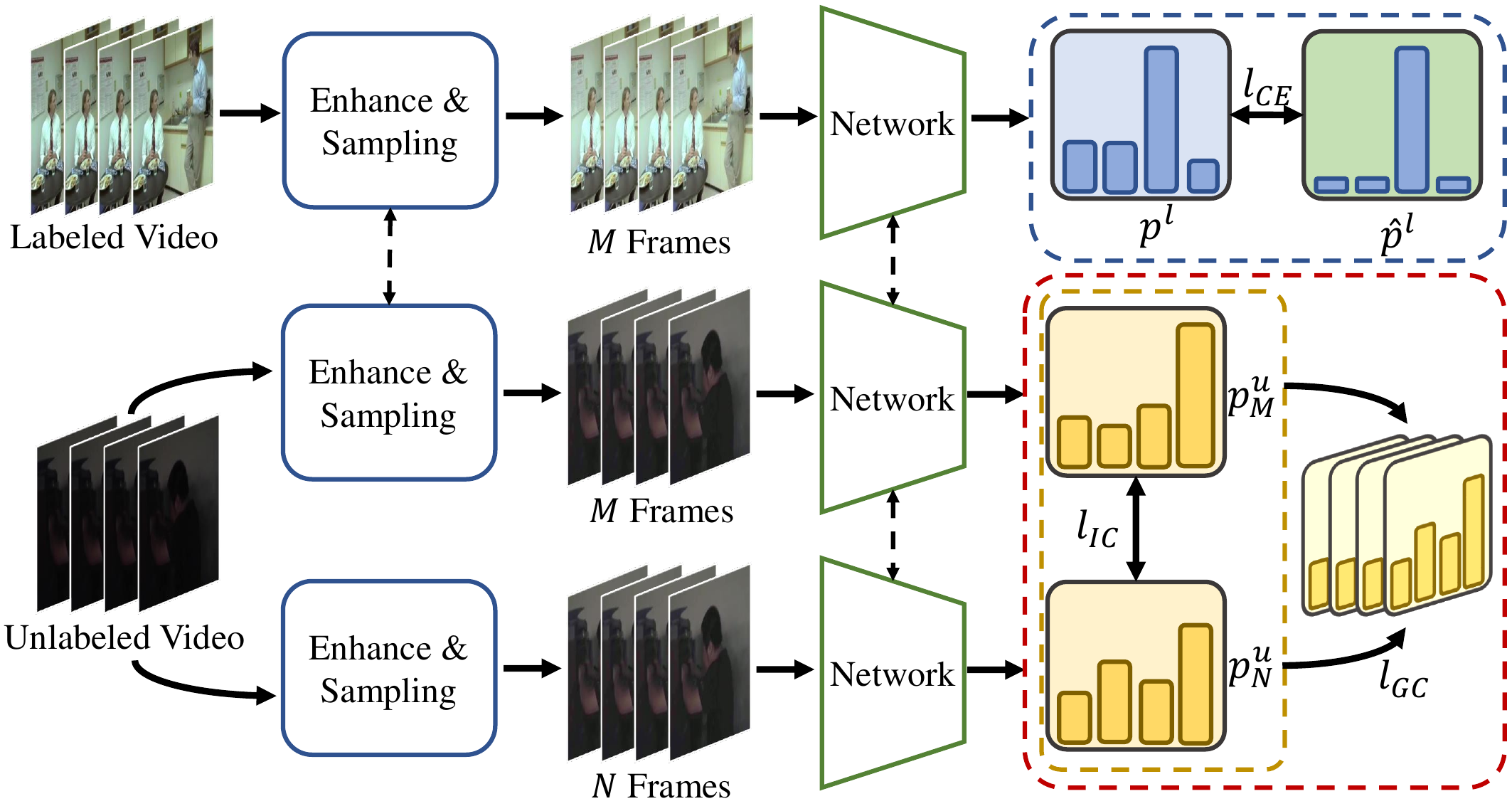}
    \label{sub-figure:4-10-2}}\\
    \subfloat[Structure of the DA-based method. Details can be found in \citep{ganin2016domain}.]{
    \includegraphics[width=0.97\linewidth]{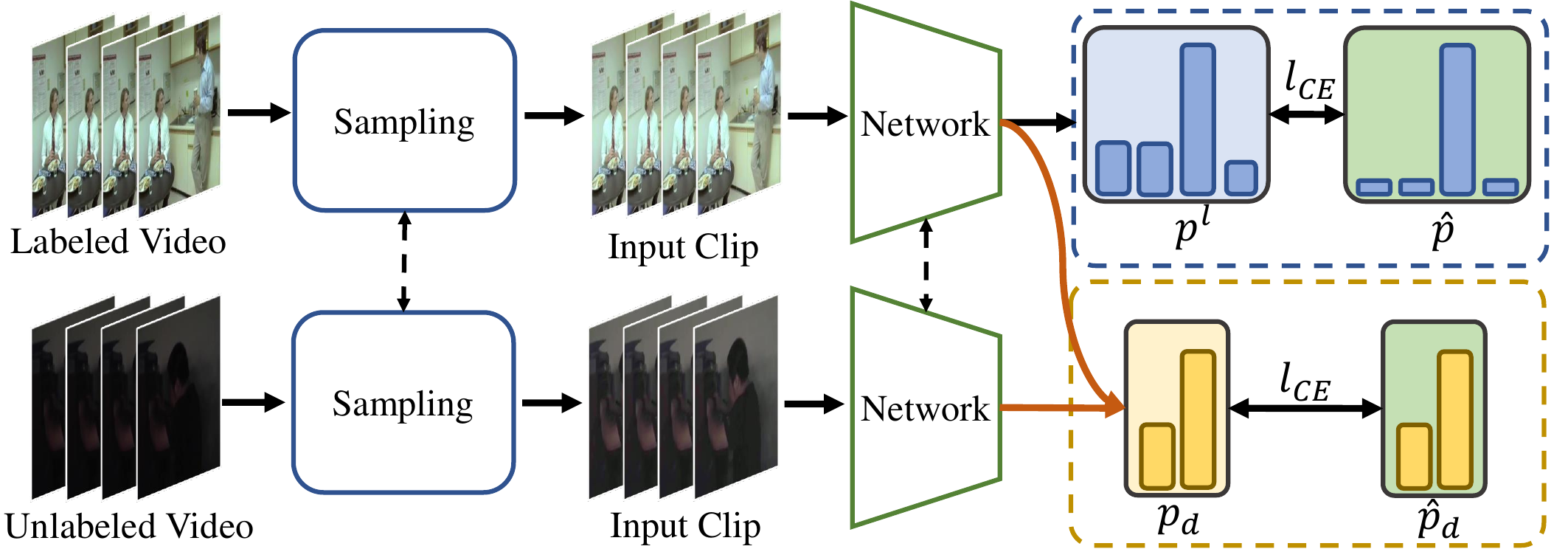}
    \label{sub-figure:4-10-3}}
    \caption{Structures utilized in UG2-2.2. Here $p$ and $\hat{p}$ indicate the prediction and the ground-truth or pseudo-label, with superscript $l$ and $u$ indicating whether it is from labeled or unlabeled sources, respectively. $l_{CE}$ is the cross entropy loss. Modules connected with dotted lines with dual arrows are identical or shared-weight. In Fig. \ref{sub-figure:4-10-2}, subscripts $M$, $N$ refer to the frame numbers of the input clips. $l_{IC}$ and $l_{GC}$ refer to Instance Contrastive (IC) loss and Group Contrastive (GC) loss, respectively. In Fig. \ref{sub-figure:4-10-3},  the $p_d$, $\hat{p}_d$ are the predictions and ground-truth for the domain classification, respectively.}
    \label{figure:4-10-2_2_methods}
\end{figure}
A total of 12 teams submitted their results in the semi-supervised challenge. Among the participants, the winner team \textit{Artificially Inspired} achieved the best performance of $93.77\%$. Similar to UG2-2.1, there is a noticeable gap between leaderboard performance and our baseline results. This is mainly because our baseline evaluation simply adopts existing domain adaptation methods without any other pre-processing or enhancement techniques to further boost the performance. Also, in order to achieve state-of-the-art performance, all teams utilized much larger epoch number (e.g. total 425 epochs for the winner team) and more complicated networks.

For the winner team \textit{Artificially Inspired}, they adopted the same backbone and enhancement method from the UG2-2.1. To fully leverage the unlabeled data from ARID, the team adopted pseudo-label strategy to create pseudo-labels for the unlabeled data as shown in Fig. \ref{sub-figure:4-10-1}. Specifically, in the first run, the team first trained the model with labeled data from HMDB51 and generated the pseudo-labels $\hat{p}^u$ of unlabeled data by inference. Samples with confident pseudo-labels were subsequently filtered based on their confidence scores and subsequently joined the supervised training process together with the data from HMDB51. The team initially chose a relatively high threshold $\sigma$ of 0.99 and further increased it up to 0.999999 from the fourth run to the tenth run. At the end of each run, the checkpoint of the trained model was saved. During the testing process, the final prediction was generated as the average of predictions from the model saved at the end of each run. 

As for the runner-up team, \textit{DeepBlueAI} achieved a competitive result of $93.63\%$ with only a minor gap of $93.77\%$ compared with the best result. The team utilized CSN \citep{tran2019video} based on ResNet-152, which is a more complex backbone compared to the winner team. While they also adopted the pseudo-label strategy similar to \textit{Artificially Inspired} as in Fig. \ref{sub-figure:4-10-1}, they designed a different set of filtering rules. Specifically, they designed a four-run training process, where all samples with pseudo-labels generated in the first run were included in the supervised training process of the second run. In the rest process, pseudo-labels of two classes, including ``Drink" and ``Pick", were changed to the class ``Pour" if satisfying one of the two following rules: (i) if the confidence score of ``Pour" is larger than 2.0, or (ii) if the confidence score of ``Pour" is larger than $1/3$ of the highest score. While the team does not reveal their rational of this design, it might be the similarity between these three actions that motivates this specific design. Also different from the winner team, \textit{DeepBlueAI} generated their prediction only based on the model of the final run. 

Other teams also provided interesting solutions to generate supervision signal from the unlabeled ARID data. For example, team \textit{Cogvengers} (rank No. 3, Top1 $84.35\%$), which utilized R(2+1)D-Bert as their based model, adopted Temporal Contrastive Learning (TCL) \citep{singh2021semi} for semi-supervised learning as shown in Fig. \ref{sub-figure:4-10-2}. Specifically, after performing GIC enhancement, the team adopted two different instance-level contrastive loss $l_{IC}$\, to maximize the mutual information between clips from the same video under different frame rates. For unlabeled samples with the same pseudo-label, a group-level contrastive loss $l_{GC}$ was utilized to minimize the feature distance within the group with the same pseudo-label. As for \textit{AstarTrek} as shown in Fig. \ref{sub-figure:4-10-3}, they adopted an adversarial-based unsupervised domain adaptation method DANN \citep{ganin2016domain} to adapt the features learned from the labeled HMDB51 data to the unlabeled data. However, they adopted a shallow backbone ResNet-18, which might be the reason for their inferior performance (Top1 $79.22\%$) compared with others.

\subsection{Analysis and Discussion}
\label{section:results:discussion}
{
\color{black}
As presented above, participants have provided various solutions to tackle action recognition in dark video for the UG2-2 challenge. All winning solutions improved substantially from the baseline results, however, there is a significant gap among the winning solutions. This justifies the difficulty of the challenge with much room for improvement.
}

In summary, advancements have been made by the various challenge solutions, all winning solutions utilize deep learning based methods with complex backbones, trained from scratch with a long training process. Such strategy possesses a high risk of overfitting given the scale of the ARID-\textit{plus}, while also suffers from the need for large computational resources. Therefore, though achieving notable performances, such strategy may not be ultimate for AR in dark videos. Meanwhile, though domain adaptation approaches have been popular in coping with semi-supervised action recognition, where dark videos are unlabeled, domain adaptation solutions are not the preeminent ones in the challenge, due to unique characteristics of dark videos. Such observation suggests that there are limitations in applying domain adaptation to dark videos directly. To further improve AR accuracy, an intuitive strategy is to apply low-light enhancement methods. However, empirical results go against such intuition. 

\textbf{Are image enhancement methods effective?} While some low-light enhancement methods do bring improvements in accuracy, results show that the improvements are erratic. Negative effects due to enhancements could be explained by its disruption over the original data distribution as well as the introduction of noise. Interestingly, the few adopted enhancements in the winning solutions may not produce the best visual enhancement results. Instead, it could be observed that these methods would either preserve the character of the original data distribution or introduce less noise. Therefore, it could be argued that for any enhancement to bring substantial improvement in AR accuracy, either condition should be met. Since less noise could contribute towards AR accuracy, employing further denoising methods \citep{tassano2020fastdvdnet,sheth2021unsupervised} could be examined along with the various low-light enhancement methods to suppress noise, mitigating the possible negative effects. Meanwhile, current solutions only exploit one single enhancement. To this end, enhancement-invariant methods may be developed to capture underlying distributions that are not influenced by enhancement methods, which could be the key to understanding dark videos. This strategy could be implemented with various enhancement methods applied simultaneously to the dark videos, with the invariant features trained by contrastive learning \citep{qian2021spatiotemporal,pan2021videomoco} of the enhanced results. The final classification would be performed on the enhancement-invariant features extracted.

\textbf{How to reduce model complexity?}
\textcolor{black}{
To overcome the risk of overfitting and the requirement for large computational resource due to the use of large-scale deep learning methods, multiple alternative strategies could be considered. One of which is to incorporate few-shot learning approaches \citep{kumar2019protogan,bo2020few}, which has enabled models to be trained with limited data while generalizing to unseen test data, and has been gaining research interest for action recognition. This conforms to the task of AR in dark environments, and should therefore be considered as a feasible alternative to the fully supervised strategy. Further, due to the insufficient number of classes in ARID-\textit{plus}, winning solutions may not be capable of generalizing to videos in the wild, where most actions are considered to be unseen by ARID-\textit{plus}. To overcome such shortcoming, zero-shot learning approaches \citep{xu2017transductive,mishra2018generative,liu2019generalized} endows AR methods the capacity of predicting unseen actions, which could better cope with real-world scenarios. Meanwhile, techniques such as self-supervised learning would also boost model capacity by exploiting extra information within videos, such as video speed \citep{wang2020self} and video coherence \citep{cao2021self}.
}
Meanwhile, to apply models in areas such as surveillance, models should be deployed on edge devices (e.g., embedding systems such as Jetson). These devices possess limited computation resources but are able to be mass deployed. These attributes prohibit large-scale models to be applied directly. One possible solution would be model compression \citep{he2018amc,pan2019compressing}, which aims to deploy models in low-power and resource-limited devices without a significant drop in accuracy. The ability of the compressed model to be applied on edge devices could help to expand the application of AR solutions in scenarios such as nighttime autonomous driving systems, where conventional hardware (i.e., GPUs and TPUs) could not be installed.

\textbf{Does domain adaptation help a lot?} Applying domain adaptation approaches directly to semi-supervised AR of dark videos is ineffective largely due to the large domain gap between normal videos and videos in dark environments. Domain adaptation approaches would therefore be unable to minimize the discrepancies between different domains, or to extract domain-invariant features for transferring. Currently, most domain adaptation approaches align high-level features \citep{ganin2016domain,ganin2015unsupervised,long2015learning,saito2018maximum}, which is in accord with the fact that high-level features are utilized for the final classification task. However, large discrepancies would exist between the low-level features of normal and dark videos, given the large differences in mean and standard deviation values of video frames. The discrepancies between low-level features would escalate the discrepancies between high-level features, therefore undermining the effort of current domain adaptation approaches in obtaining transferable features from normal videos. In view of such observation, low-level features should be aligned with high-level features jointly when designing domain adaption approaches for semi-supervised AR in dark videos.
    
\textbf{How to leverage multi-modality information?} Besides the techniques mentioned above, it is observed that optical flow could bring performance improvement. Optical flow can be viewed as an additional modality embedded in videos, and could provide more effective information thanks to the fact that it is essentially computed as the correlation of spatiotemporal pixels between successive frames, which is highly related to motion. However, in solutions utilizing optical flow, it is extracted with hand-crafted methods, such as TVL1, which require a large computation cost. Hand-crafted optical flow also prohibits end-to-end training due to the need for storing optical flow before subsequent training. Advances have been made in optical flow estimation with deep learning method \citep{ranjan2017optical,hui2018liteflownet,sun2019models} that allows optical flow estimation to be performed along with the training of feature extractors and classifiers in an end-to-end manner. However, these advances are made with normal illuminated videos, and it is worth exploring whether these models could also be applied with videos shot in dark environments.  Meanwhile, with the optical flow as an additional modality of information, current solutions tend to utilize optical flow independent from RGB features, with the results obtained in a late fusion fashion. Since both modalities are obtained from the same set of data, it would be worth exploring how to train with both modalities jointly through approaches such as cross modality self-supervision \citep{khowaja2020hybrid,sayed2018cross}, which can be applied in both supervised training and cross-domain semi-supervised training \citep{munro2020multi}. Such approach enables network to learn features with high semantic meaning, which could lead to further improvements in AR effectiveness.

{
\color{black}
\subsection{Future Work}
\label{section:results:future}
The introduction of the UG\textsuperscript{2}+ Challenge Track 2 and its two sub-challenges promotes research in both fully supervised and semi-supervised settings for action recognition in dark videos, and have received various solutions. However, it is observed that the best solutions are tailored to the sub-challenges, which is characterized by its limited data with no limits on training resource. The best solutions therefore tend to possess long training process, large batch size and complex structure, which may not be applicable in resource-constrained real-world scenarios. In future workshops, we will explore on more realistic dark video action recognition by taking resource constraints into account and by increasing the complexity of the dataset. Meanwhile, the current sub-challenges are only designed for fully or semi-supervised learning for action recognition, and have not considered the settings of either few-shot learning or zero-shot learning, which are all more realistic settings given that it is not always applicable to gain sufficient training data in real-world applications. We would therefore open up new tracks that explore few-shot learning and zero-shot learning settings for AR in dark videos in future workshops.
}

\section{Conclusion}
\label{section:concl}

In this work, we dive deeper into the challenging yet under-explored task of action recognition (AR) in dark videos, with the introduction of a novel UG\textsuperscript{2}+ Challenge Track 2 (UG2-2). UG2-2 aims to promote the research of AR in challenging dark environments from both fully supervised and semi-supervised manners, improving the generability of AR models in dark environments. Our baseline analysis justifies the difficulties of the challenges, with poor results obtained from current AR models, enhancement methods and domain adaptation methods. While solutions in UG2-2 has introduced promising progress, there remain large room for improvements. We hope this challenge and the current progress could draw more interest from the community to tackle AR in dark environments.

\section*{Declarations}

\begin{itemize}
\item Funding: No funding was received to assist with the preparation of this manuscript.
\item Competing interests: The authors have no competing interests to declare that are relevant to the content of this article.
\item Data availability: The datasets generated during and/or analysed during the current study are available in the ARID repository, \url{http://xuyu0010.github.io/arid.html}
\end{itemize}






\clearpage
\bibliography{arid_ijcv}

\end{document}